\documentclass[runningheads]{llncs}
\usepackage[T1]{fontenc}
\usepackage{amsmath}
\usepackage{float}
\usepackage{amsfonts}
\usepackage{amssymb}
\usepackage{graphicx}
\usepackage{subcaption}
\begin{document}
\title{Longitudinal Mammogram Exam-based Breast Cancer Diagnosis Models: Vulnerability to Adversarial Attacks }
%
%

\author{{Zhengbo Zhou$^{\star}$  \qquad  Degan Hao$^{\star}$ \qquad  Dooman Arefan$^{\dagger}$  \qquad  Margarita Zuley$^{\dagger}$        \\     \qquad  Jules Sumkin$^{\dagger}$ \qquad Shandong Wu$^{\star \dagger \mathsection}$}}
%

%
\institute{{$^{\star}$Intelligent Systems Program, University of Pittsburgh, Pittsburgh, PA, USA\\
    $^{\dagger}$Department of Radiology, University of Pittsburgh, Pittsburgh, PA, USA\\
    $^{\mathsection}$Department of Biomedical Informatics and Department of Bioengineering\\University of Pittsburgh, Pittsburgh, PA, USA}}

\maketitle              
\begin{abstract}
In breast cancer detection and diagnosis, the longitudinal analysis of mammogram images is crucial. Contemporary models excel in detecting temporal imaging feature changes, thus enhancing the learning process over sequential imaging exams. Yet, the resilience of these longitudinal models against adversarial attacks remains underexplored. In this study, we proposed a novel attack method that capitalizes on the feature-level relationship between two sequential mammogram exams of a longitudinal model, guided by both cross-entropy loss and distance metric learning, to achieve significant attack efficacy, as implemented using attack transferring in a black-box attacking manner. We performed experiments on a cohort of 590 breast cancer patients (each has two sequential mammogram exams) in a case-control setting. Results showed that our proposed method surpassed several state-of-the-art adversarial attacks in fooling the diagnosis models to give opposite outputs. Our method remained effective even if the model was trained with the common defending method of adversarial training. 

\end{abstract}

\section{Introduction}
Mammography-based AI is at the forefront of medical AI research and many AI products are being translated for clinical deployment. The cyber-security of such models is therefore becoming a paramount need to ensure AI integrity. Although medical IT system is relatively closed, but cyber-attacks still occur, quite often, through infiltration of the IT systems \cite{cite1} or by internal hackers \cite{mirsky2019ct}. For example, hospitals had to pay ransom when their data are hacked \cite{cite3}. In fact, 94 \% of U.S. hospitals are affected by healthcare data breaches \cite{cite5}, showing high vulnerability and risks to adversarial attacks. As elucidated in \cite{finlayson2019adversarial}, adversarial attacks can occur with real world bad intentions for data-based ransom, insurance fraud, clinical trial effect manipulation, etc.

In clinical radiology, the established practice of comparing prior data with current data serves as a cornerstone for radiologists to perform lesion detection and diagnosis. The emergence of deep learning models has refined this process by integrating temporal sequential data, such as using multiple mammograms, magnetic resonance imaging, computed tomography, etc., as model inputs. Recent evidences indicate that these longitudinal models outperform those relying solely on a single time-point input. Cui et al. \cite{cui2019rnn} shows that applying convolutional neural networks to follow-up MRI scans can effectively diagnose Alzheimer's disease. Dadsetan et al. \cite{dadsetan2022deep} proposes a novel deep learning model on longitudinal mammogram showing superior effects  compared to the model using a single exam. Lee et al. \cite{lee2023enhancing}
leverages a Transformer decoder to incorporate prior images showing improved performance on breast cancer risk prediction.

In the biomedical domain, longitudinal models using sequential imaging scans are gaining popularity, where these models are effective by utilizing spatiotemporal relationships of longitudinal data. Adversarial attacks on models using an individual time-point scans have received considerable attention \cite{zhou2021machine}. However, there has been little to none exploration into attacks on longitudinal models.

In this work, we focus on the task of diagnosing breast cancer through a Transformer decoder architecture \cite{vaswani2017attention} using sequential mammogram exams. We examine the susceptibility of this longitudinal model to adversarial attacks. We propose a novel attack method tailored to combine cross-entropy loss (to guide adversarial samples across the decision boundary) and distance metric learning (to modify the relationship between sequential imaging exams), aiming to fool the diagnosis models to give rise to an opposite output. We followed the attack transferring scheme and showed through experiments that, even when adversarial training is employed to enhance the model’s robustness, our attacking method can still effectively degrade the model’s performance. Our method also outperforms several other compared methods. Our study highlights a significant vulnerability of longitudinal models to adversarial attacks, which urges the needs of enhancing such model’s safety against adversarial attacks.

The contributions of our work can be summarized as: (1) We studied a novel topic of investigating adversarial attacks on longitudinal imaging-based diagnostic models; (2) We evaluated the method of integrating cross-entropy loss with distance metric learning to implement the attack, which exhibited stronger attaching effects in leading to misclassification of breast cancer, compared to several state-of-the-art attack techniques; and (3) We showed that in distance metric learning, medical knowledge can be leveraged in selecting effective adversarial samples aiming to fool a diagnosis model.

\section{Related Work}

\noindent \textbf{Adversarial Attack Methods: } Szegedy et al \cite{szegedy2013intriguing} showed that classifiers may confidently make incorrect predictions when subjected to imperceptible perturbations. Kurakin et al. \cite{kurakin2016adversarial} showed that classifiers remain vulnerable to adversarial samples even in the physical world. Goodfellow et al. \cite{goodfellow2014explaining} propose FGSM, a simple and efficient adversarial samples generating method. In addition, Basic Iterative Method (BIM) \cite{kurakin2018adversarial} is proposed as an extension of the Fast Gradient Sign Method (FGSM) and aims to generate stronger adversarial samples by iteratively applying FGSM with smaller perturbation steps. Benefiting from the iteration strategy, MI-FGSM is proposed by adding momentum through iterations \cite{dong2018boosting}. The Projected Gradient Descent (PGD) attack \cite{madry2017towards} is another popular and powerful attack method. On the other hand, optimization-based  C\&W attack \cite{carlini2017towards} treat the adversarial sample generation as an optimization problem. 

\noindent \textbf{Attack Transferability: } The vulnerability of models to adversarial samples generated by other models is known as the transferability of adversarial attacks \cite{szegedy2013intriguing}. Kurakin et al. \cite{kurakin2016adversarial} conducted a study on adversarial samples within the context of ImageNet. Their found that BIM’s multi-step approach is less transferable than the single-step FGSM. Zhou et al. \cite{zhou2018transferable} showed that enhancing the transferability of the BIM can be achieved by maximizing the distance between natural images and their adversarial counterparts within the intermediate feature maps, coupled with the addition of regularization. The transferability of a single-input model to longitudinal models is an interesting approach but not yet studied in the literature.

\noindent \textbf{Defense Methods:} Hinton et al. \cite{hinton2015distilling} introduced distillation as a method to bolster resistance against adversarial samples. The adversarial training method involves incorporating adversarial samples into the training set to enhance model robustness through retraining \cite{kurakin2016adversarial}. Tramèr et al. \cite{tramer2017ensemble} expanded on this approach with ensemble adversarial training, augmenting training data with perturbations transferred from other models. In general, adversarial training is a common defending method for adversarial attacks. 

\begin{figure}
\includegraphics[width=\textwidth]{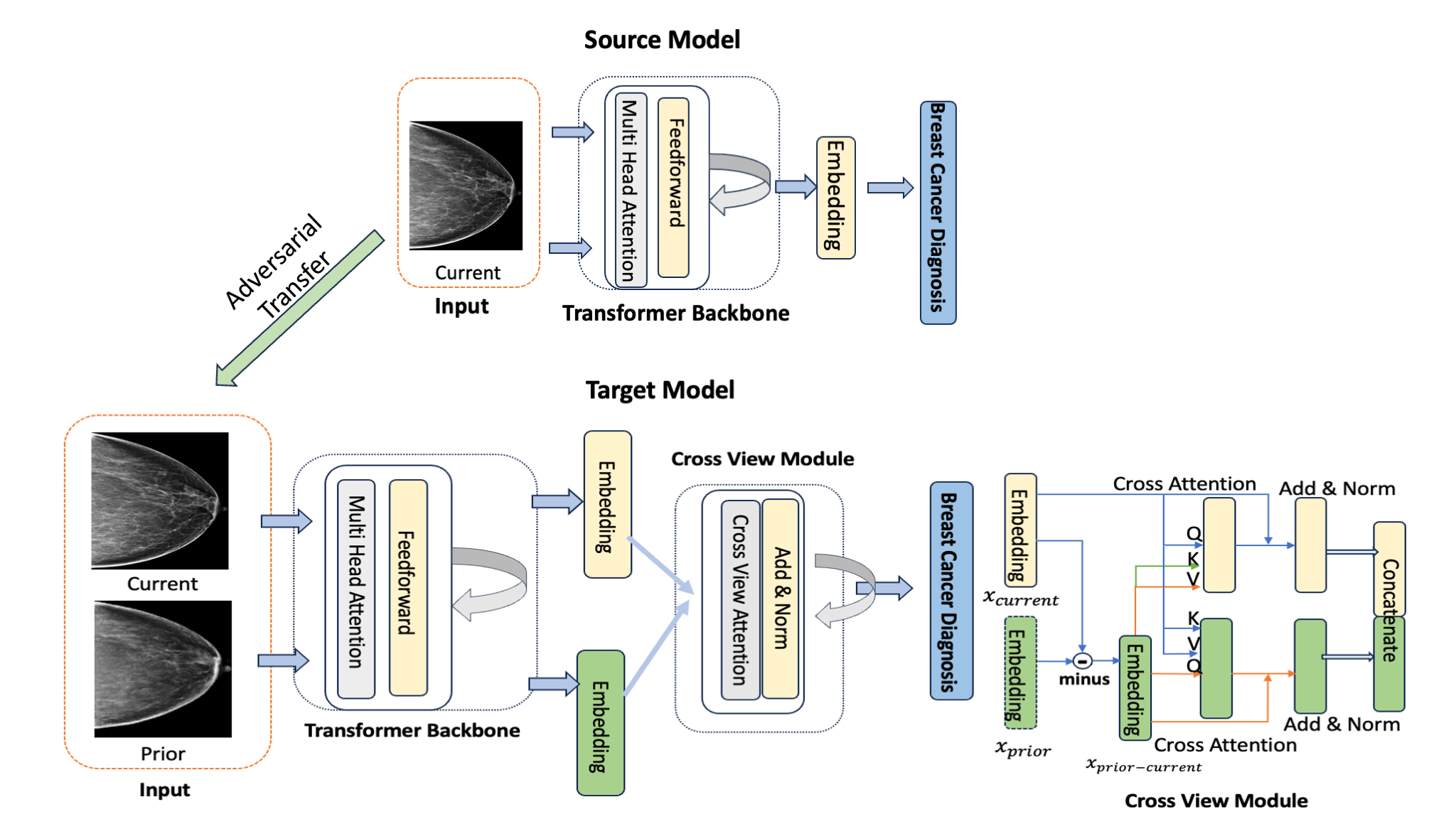}

\caption{Adversarial attacks transferred from a source model (a single input, Current) to a target model (two sequential input, Prior and Current) for breast cancer diagnosis.} \label{fig1}
\vspace{-2.0em}
\end{figure}

\section{Methods}

\subsection{Task and Model Architecture}
\noindent \textbf{Task.} In this study, we implemented our novel method on the task of diagnosing breast cancer on mammograms (i.e., classification of cancer cases vs. normal controls). The diagnosis can be based on a single time-point exam or multiple longitudinal exams that leverage temporal information to improve diagnosis accuracy. Thus, we designed two models: Source model (using a single time-point mammogram exam, denoted as Current) and Target model (using two sequential exams, denoted as Prior and Current, respectively, which are taken at two different time points). The Source model is useful when a patient does not have Priors available. It should be pointed out that the Prior exams are normal/healthy mammograms that are stored as historical data of a patient in the electronic health records. The use of normal  Priors in longitudinal models conforms to radiologists’ clinical practice to improve lesion detection and diagnosis in addition to using the Current exam. 

\noindent \textbf{Model Architecture.} As illustrated in Figure 1, in the Source model, features extracted by the backbone model are utilized for diagnosis, which also serve as the basis for generating adversarial samples to attack the target longitudinal model i.e., (the attack transferring). 
For the Target longitudinal model, the imaging features are derived from both the Prior and Current exams, referred to as $x_{\text{prior}}$ and $x_{\text{current}}$, respectively. The inputs to the cross-view module are current exam $x_{\text{current}}$ and the changes between the prior and current exam, i.e., $x_{\text{prior-current}} = x_{\text{prior}} - x_{\text{current}}$. Here  $x_{\text{prior-current}}$ aims to capture and emphasize the changes between the two time points. Subsequently, both feature vectors are fed to the cross-view module. In this module, we use multi-head attention mechanisms to identify and stress the information in the subtracted feature ($x_{\text{prior-current}}$) that is relevant to the current feature ($x_{\text{current}}$). The core of multi-head attention is a scaled dot-product attention\cite{vaswani2017attention} mechanism that computes attention scores between queries and keys, which are then used to linearly combine the associated values. In our setting, we reshape the embedded feature maps for the target phase ($x_{\text{current}}$ or $x_{\text{prior-current}}$) into a query matrix and reshape the feature maps for the source phase into a key matrix.
The outcome of this attention process produces  \(x_{\text{current-attention}}\) and \(x_{\text{(prior-current)-attention}}\).  To form new attention-enhanced feature vectors, we integrate these representations with the original vectors, resulting in $x_{\text{(prior-current)-cross}} = x_{\text{prior-current}} + x_{\text{(prior-current)-attention}}$ and $
x_{\text{current-cross}} = x_{\text{current}} + x_{\text{current-attention}}$. These refined feature vectors are then concatenated, serving for the ultimate prediction.
\subsection{Adversarial Attack}

\textbf{Rationale:} Our study aims to attack the Target model that makes diagnosis using the temporal relationships/changes of the Prior and Current exams. In our black-box attacking scenario, attackers do not have access to the model architecture and parameters of the Target model, particularly the mechanisms in which how the longitudinal features are used. Attackers craft adversarial samples using the Source model, following the scheme of “attack transferring”, to perform attacks to the Target model. This makes sense because attackers do not need to or may not have a longitudinal models to generate adversarial samples; instead, a Source model based on a single time-point input is much more accessible and easily available to use. Here, the attacks aim to manipulate the Current exam without considering the Prior exam, as the cancer diagnosis is primarily based on the Current exam (recall that the Prior exams are all normal/healthy and they serve as comparisons to Current exams to improve cancer detection).

\begin{figure}
\vspace{-4.0em}
\includegraphics[width=\textwidth]{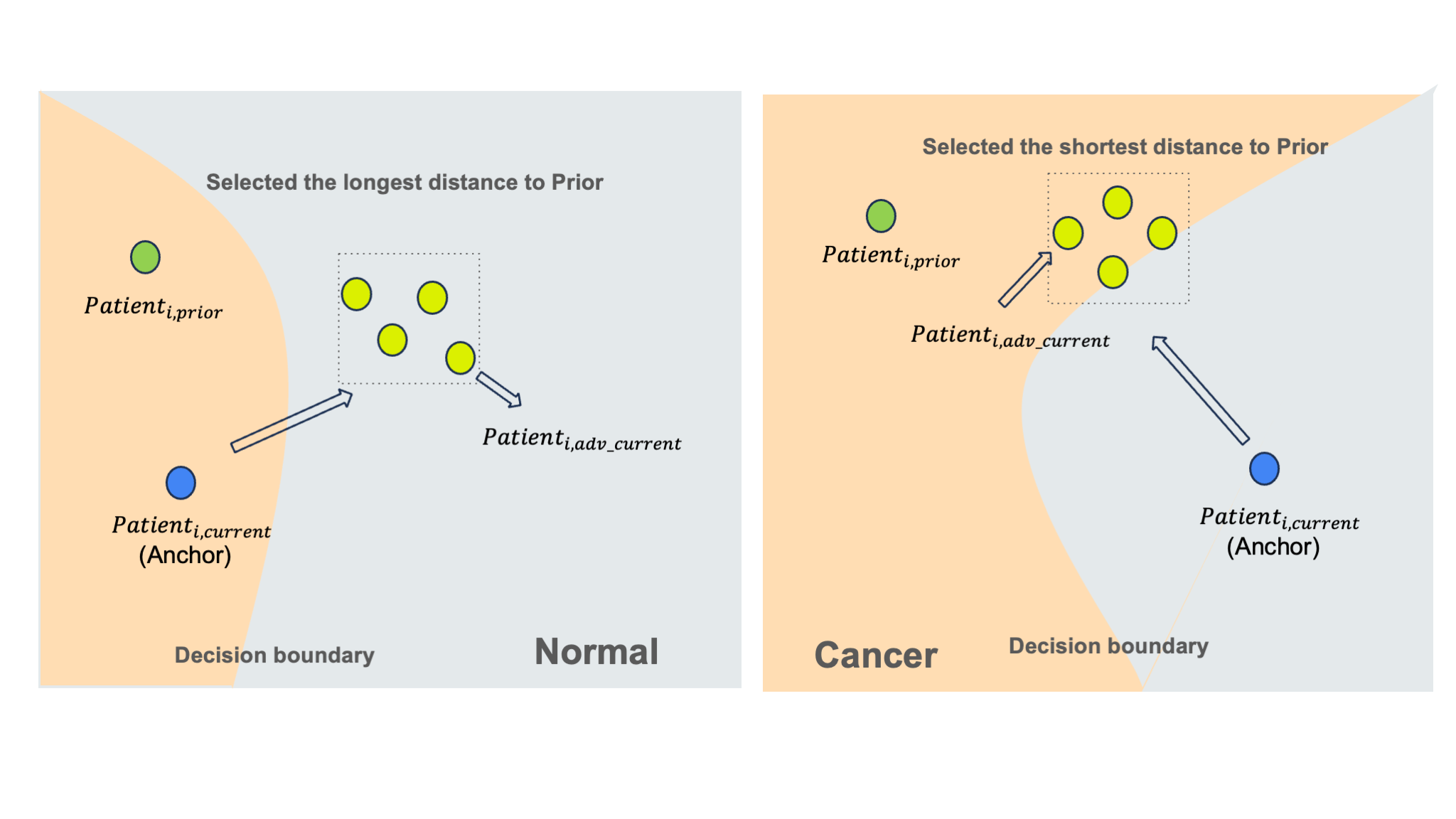}
\vspace{-5.0em}
\caption{Illustration of knowledge-guided selection of samples for adversarial Current: to select the one with the shortest (longest) distance to Prior for Cancer (Control) cases. Light orange represents Normal decision space, while gray represents Cancer decision space.} \label{fig1}
\vspace{-1.0em}
\end{figure}

\noindent \textbf{Adversarial Sample Generation}. Let \( f(s) \) denote an arbitrary deep neural network which takes \( s \) (\( s \in \mathbb{R}^n \)) as input and outputs the probability of classes \( y \) (\( y \in \mathbb{R}^m \)). We first define an adversarial sample fooling the model \( f(s) \) for a chosen \( p \)-norm and noise parameter \( \epsilon \) as follows:
\begin{equation}
\tilde{s} = \text{argmax}_{\| \tilde{s} - s \|_p \leq \epsilon} l(\tilde{s}, t)
\end{equation}
where \( t \) and \( l(\cdot, \cdot) \) denote the label of \( x \) and the loss function used to train the model respectively. In our experiments, we investigate the effects of using the cross entropy loss and distance metric learning for attacking the Target model. To optimize Equation (1), the Iterative Fast Gradient Sign Method (I-FGSM) is employed. Initially, the input \( x \) is scaled to the range into \([-1,1]\) and we set \( \tilde{x}_0 = x \). Then, we compute the gradients of loss function with respect to input \( x \). Following this, the adversarial samples undergo several iterations of updates. In each step, we apply the sign function to the calculated gradients and then clip the updated adversarial samples to the range \([-1,1]\) to maintain them as valid images. At every iteration, the generated samples are designed to cross the decision boundary, directed by the gradient data, and are retained for subsequent sampling. 
Ultimately, adversarial samples are generated by adding the pixel differences, scaled by 
\( \epsilon \), between the most recently updated adversarial samples and the original input \( x \).

\noindent \textbf{Knowledge-guided Adversarial Sample Selection Method}. As illustrated in Figure 2, the I-FGSM method can generate multiple adversarial samples aiming at traversing the decision boundary. We propose a novel criteria to select most impacting adversarial samples using distance metric learning. This criteria is designed based on the knowledge of: (1) the longitudinal model relies on the relationship of Prior (always normal) and Current (could be cancer or normal) exams to make a diagnosis, and (2) the Euclidean distance between Prior and Current is smaller for a normal patient (because of less variation across two normal exams), but larger for a cancer patient (because of larger variation from normal transitioning to cancer development). Specifically the criteria differ between Normal and Cancer cases: in Cancer cases, we select the adversarial sample $Patient_{i,adv\_current}$ that is closest from $Patient_{i,prior}$, while in Control cases, we choose the sample that is furthest to $Patient_{i,prior}$. For Cancer cases, the objective is to coax the model into misclassifying the adversarial sample as Normal by leveraging the proximity to a $Patient_{i,prior}$ image labeled as normal. This proximity strategy enhances the likelihood that the longitudinal model, which relies on temporal relationships for classification, erroneously assesses the sample as Normal. Conversely, for Normal cases, the intent behind selecting adversarial samples with the greatest distance from the Prior image is to challenge the model into a false diagnosis of Cancer. Given the substantial separation between the adversarial sample and a Prior image, the longitudinal model is more inclined to misclassify these as belonging to Normal cases. This novel adversarial sample selection approach takes advantages of the longitudinal model’s dependencies on the relationship between $Patient_{i,prior}$ and $Patient_{i,current}$ exams.

\section{Experiments}
\subsection{Study Cohort}
This study received institutional review board approval and we used a dataset of 590 subjects in a case-control study setting, with 293 breast cancer cases and 297 breast cancer-free controls (i.e., normal/negative). Each subject has a Current mammogram exam and a Prior exam taken at approximately 1 year apart. All diagnosis outcomes are biopsy-proven and based on the Current exams. All Prior exams are normal/negative. For the cancer cases, we exclusively utilized the biopsied breast for. For the controls, either the right or left breast was randomly chosen for a subject to avoid modeling from shortcut learning. To ensure uniformity in the orientation of a breast, we applied a horizontal flip for a right-side breast to appear like a left-side breast. We specifically opted for using the craniocaudal view of the mammogram images. 

\subsection{Implementation Details}
The dimensions of the input mammogram images were standardized to 350x400 pixels to maintain consistency across the subjects. We employed a Swin Transformer \cite{liu2021swin} pretrained on ImageNet as the backbone for feature extraction in both the Source and Target models. We also evaluated the effects of implementing the Source model with a VGG \cite{simonyan2014very} model as the backbone architecture, whereas the Target model remains unchangedand and reported results in Supplementary materials. To prevent overfitting for both Source and Target model training, we incorporated data augmentation during the training phase, including flipping and image rotation. The model underwent training over 30 epochs, with learning rates of 5e-5 and 1e-5 identified as the optimal parameter values through experiments. For generating adversarial attacks, the iteration number was set to 15 for all iterative-based attack methods, and the perturbation size was set to 0.01 for all attacks. We performed parameter robustness analysis experiments on these two parameters (See results in Supplementary materials).\\
We compared the effects of our method to several methods, including C\&W, FGSM, I-FGSM, MI-FGSM, and PGD. Also, we compared to two distance-guided new methods using distance loss on FGSM and I-FGSM, respectively, as described by Equation (2), where we denote x as input data, t as label, L(x) as the intermediate feature map:

\vspace{-1.0em}
\begin{equation}
\text{loss}_{\text{distance}}(x', t) = 
\begin{cases} 
L(x') - L(x_{\text{prior}}) & \text{if } t = 1 \\
L(x_{\text{prior}}) - L(x') & \text{if } t = 0
\end{cases}
\end{equation}
In addition, we implemented two more novel methods that used distance metric loss to alter the relationship between Prior and adversarial Current. Specifically, we use the distance loss in Equation (2) as a penalty term aiming to more efficiently guide the search directions. We named this regularization method as 'Distance Reg.' (Equation (3)) as another way to combine the cross entropy loss and distance metric learning. We compared our proposed knowledge-guided method to this alternative method ($\lambda$ is experimentally determined as 0.05).
\begin{equation}
\tilde{s} = \text{argmax}_{\| \tilde{s} - s \|_p \leq \epsilon} l(\tilde{s}, t) + \lambda \cdot \text{loss}_{\text{distance}}(s', t)
\end{equation}
We utilized a patient-wise 5-fold cross-validation and Area Under the ROC Curve (AUC) reporting Mean AUC ± Standard Deviation (Std) of the various methods. We also assessed the model’s performance after incorporating adversarial training as defence method, where the model underwent retraining with a combination of clean data and adversarial samples generated by BIM with perturbation size of 0.01 and batch size of 32. All computational tasks were performed on an NVIDIA TESLA V100 GPU, provided by our local supercomputing facility.

\begin{table}[H]
\vspace{-2em}
\centering
\caption{The performance of different attacks on Source, Target without, and Target with adversarial training models. (Format: mean AUC ± std).}
\begin{tabular}{lccc}
\hline
Attack Method                                         & Source Model                          & Target Model                 & \begin{tabular}[c]{@{}c@{}}Target Model\\ (Adversarial Training)\end{tabular} \\ \hline
No Adversarial Attack                                 & $0.670 \pm 0.017$                     & $0.704 \pm 0.054$            & $0.685 \pm 0.033$                                                             \\
FGSM \cite{goodfellow2014explaining} & $0.194 \pm 0.052$                     & $0.405 \pm 0.045$            & $0.584 \pm 0.044$                                                             \\
Distance-guided FGSM                                  & $0.403 \pm 0.063$                     & $0.531 \pm 0.040 $           & $0.634 \pm 0.029$                                                             \\
Distance Reg. FGSM                                    & \multicolumn{1}{l}{$0.292 \pm 0.080$} & $0.452 \pm 0.053 $           & $0.606 \pm 0.031$                                                             \\
I-FGSM \cite{kurakin2018adversarial} & $0.0 \pm 0.0$                         & $0.322 \pm 0.056 $           & $0.573 \pm 0.038$                                                             \\
Distance-guided I-FGSM                                & $0.039 \pm 0.014$                     & $0.430 \pm 0.030 $           & $0.612 \pm 0.037$                                                             \\
Distance Reg. I-FGSM                                  & $0.0 \pm 0.0 $                        & $0.286 \pm 0.049 $           & $0.573 \pm 0.031$                                                             \\
C\&W \cite{carlini2017towards}       & $0.0 \pm 0.0 $                        & $0.376 \pm 0.063$            & $0.572 \pm 0.035$                                                             \\
MI-FGSM \cite{dong2018boosting}      & $0.0 \pm 0.0 $                        & $0.297 \pm 0.046$            & $0.570 \pm 0.043$                                                             \\
PGD \cite{madry2017towards}          & $0.0 \pm 0.0 $                        & $0.289 \pm 0.064$            & $0.573 \pm 0.031$                                                             \\
AutoAttack   \cite{croce2020reliable}                                         & $0.0 \pm 0.0 $                        & $0.308 \pm 0.042$            & $0.585 \pm 0.034$                                                             \\
GI-FGSM     \cite{wang2022boosting}                                          & $0.0 \pm 0.0 $                        & $0.286 \pm 0.04$             & $0.587 \pm 0.028$                                                             \\
PC-I-FGSM     \cite{wan2023adversarial}                                        & $0.0 \pm 0.0 $                        & $0.291 \pm 0.05$             & $0.584 \pm 0.016$                                                             \\
GRA      \cite{zhu2023boosting}                                             & $0.0 \pm 0.0 $                        & $0.303 \pm 0.045$            & $0.574 \pm 0.036$                                                             \\
\textbf{Our Proposed Attack}                          & $0.0 \pm 0.0 $                        & $ \textbf{0.205} \pm 0.040 $ & $ \textbf{0.548} \pm 0.036 $                                                  \\ \hline
\end{tabular}
\end{table}

\begin{table}[H]
\vspace{-4em}
\caption{The performance of different attack methods when using VGG as the backbone for the Source model. The Target models (including the one with adversarial training) remained the same (i.e., Swin Transformer). (Format: mean AUC ± std). }
\begin{tabular}{lccc}
\hline
Attack Method          & \begin{tabular}[c]{@{}c@{}}Source Model\\ VGG\end{tabular} & \begin{tabular}[c]{@{}c@{}}Target Model\\ Swin Transformer\end{tabular} & \begin{tabular}[c]{@{}c@{}}Target Model\\ (Adversarial Training)\end{tabular} \\ \hline
No Adversarial Attack  & $0.66 \pm 0.045$                                           & $0.704 \pm 0.054$                                                       & $0.685 \pm 0.033$                                                             \\
FGSM     \cite{goodfellow2014explaining}              & $0.161 \pm 0.055$                                          & $0.546 \pm 0.05$                                                        & $0.597 \pm 0.039$                                                             \\
Distance-guided FGSM   & $0.365 \pm 0.043$                                          & $0.545 \pm 0.045 $                                                      & $0.615 \pm 0.029$                                                             \\
Distance Reg FGSM      & \multicolumn{1}{l}{$0.292 \pm 0.08$}                       & $0.452 \pm 0.053 $                                                      & $0.606 \pm 0.031$                                                             \\
I-FGSM    \cite{kurakin2018adversarial}             & $0.0 \pm 0.0$                                              & $0.45 \pm 0.078 $                                                       & $0.595 \pm 0.038$                                                             \\
Distance-guided I-FGSM & $0.0 \pm 0.0$                                              & $0.515 \pm 0.079 $                                                      & $0.603 \pm 0.037$                                                             \\
Distance Reg I-FGSM    & $0.0 \pm 0.0 $                                             & $0.447 \pm 0.049 $                                                      & $0.596 \pm 0.031$                                                             \\
C\&W    \cite{carlini2017towards}                & $0.0 \pm 0.0 $                                             & $0.502 \pm 0.074$                                                       & $0.580 \pm 0.035$                                                             \\
MI-FGSM   \cite{dong2018boosting}             & $0.0 \pm 0.0 $                                             & $0.446 \pm 0.082$                                                       & $0.591 \pm 0.043$                                                             \\
PGD      \cite{madry2017towards}              & $0.0 \pm 0.0 $                                             & $0.461 \pm 0.113$                                                       & $0.596 \pm 0.017$                                                             \\
AutoAttack     \cite{croce2020reliable}        & $0.0 \pm 0.0 $                                             & $0.514 \pm 0.066$                                                       & $0.585 \pm 0.019$                                                             \\
GI-FGSM      \cite{wang2022boosting}           & $0.0 \pm 0.0 $                                             & $0.475 \pm 0.088$                                                       & $0.617 \pm 0.011$                                                             \\
PC-I-FGSM      \cite{wan2023adversarial}         & $0.0 \pm 0.0 $                                             & $0.53 \pm 0.071$                                                        & $0.626 \pm 0.012$                                                             \\
GRA      \cite{zhu2023boosting}                  & $0.0 \pm 0.0 $                                             & $0.464 \pm 0.115$                                                       & $0.614 \pm 0.016$                                                             \\
Proposed Attack        & $0.0 \pm 0.0 $                                             & $ \textbf{0.418} \pm 0.1 $                                              & $ \textbf{0.577 } \pm 0.016 $                                                 \\ \hline
\end{tabular}
\end{table}
\begin{figure}[ht]

\begin{subfigure}[a]{0.48\textwidth}
    \includegraphics[width=\textwidth]{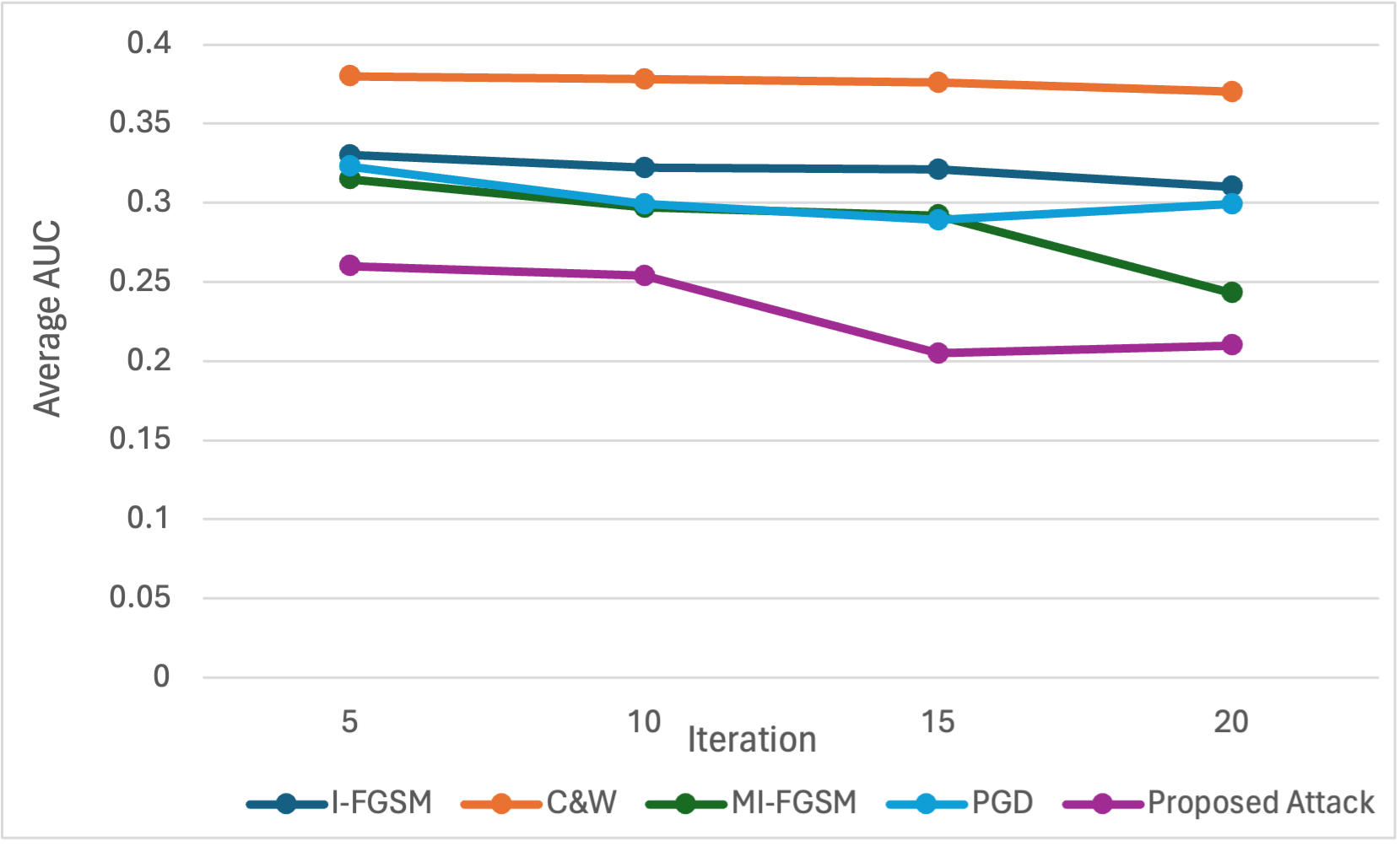}
    \caption{}
    \label{fig:sub1}
\end{subfigure}
\begin{subfigure}[a]{0.48\textwidth}
    \includegraphics[width=\textwidth]{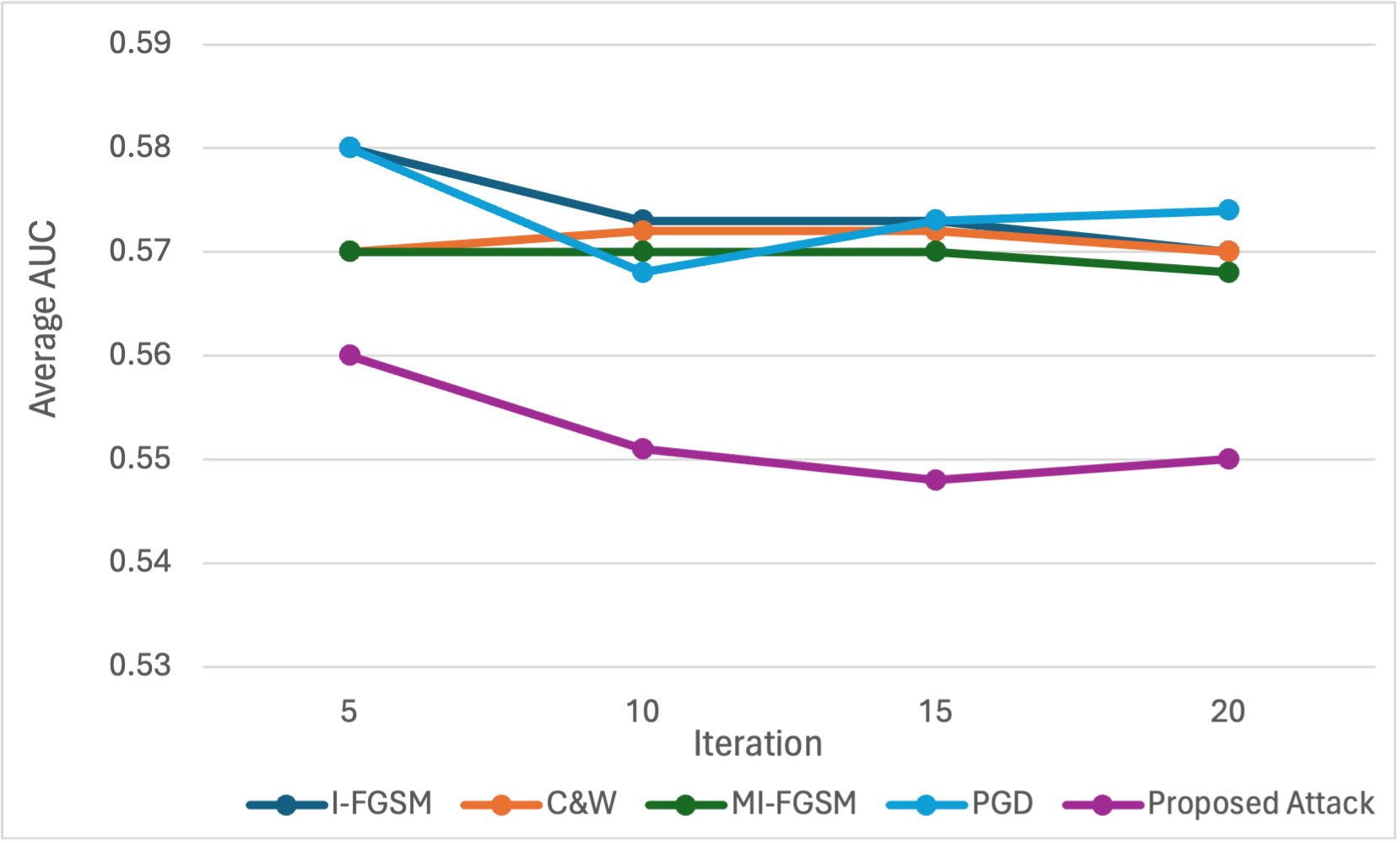}
    \caption{}
    \label{fig:sub1}
\end{subfigure}
\begin{subfigure}[a]{0.48\textwidth}
    \includegraphics[width=\textwidth]{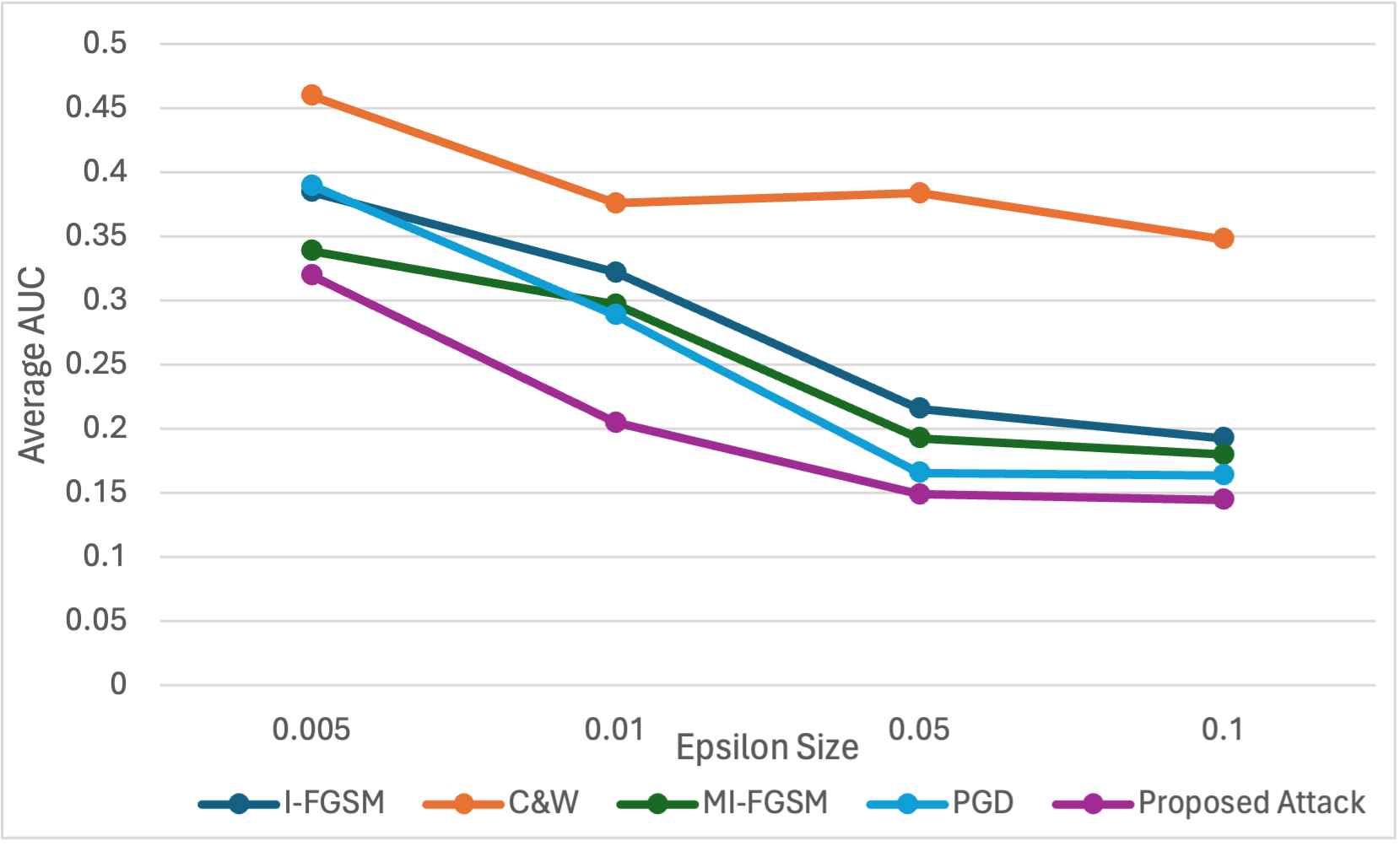}
    \caption{}
    \label{fig:sub1}
\end{subfigure}
\hfill
\begin{subfigure}[a]{0.48\textwidth}
    \includegraphics[width=\textwidth]{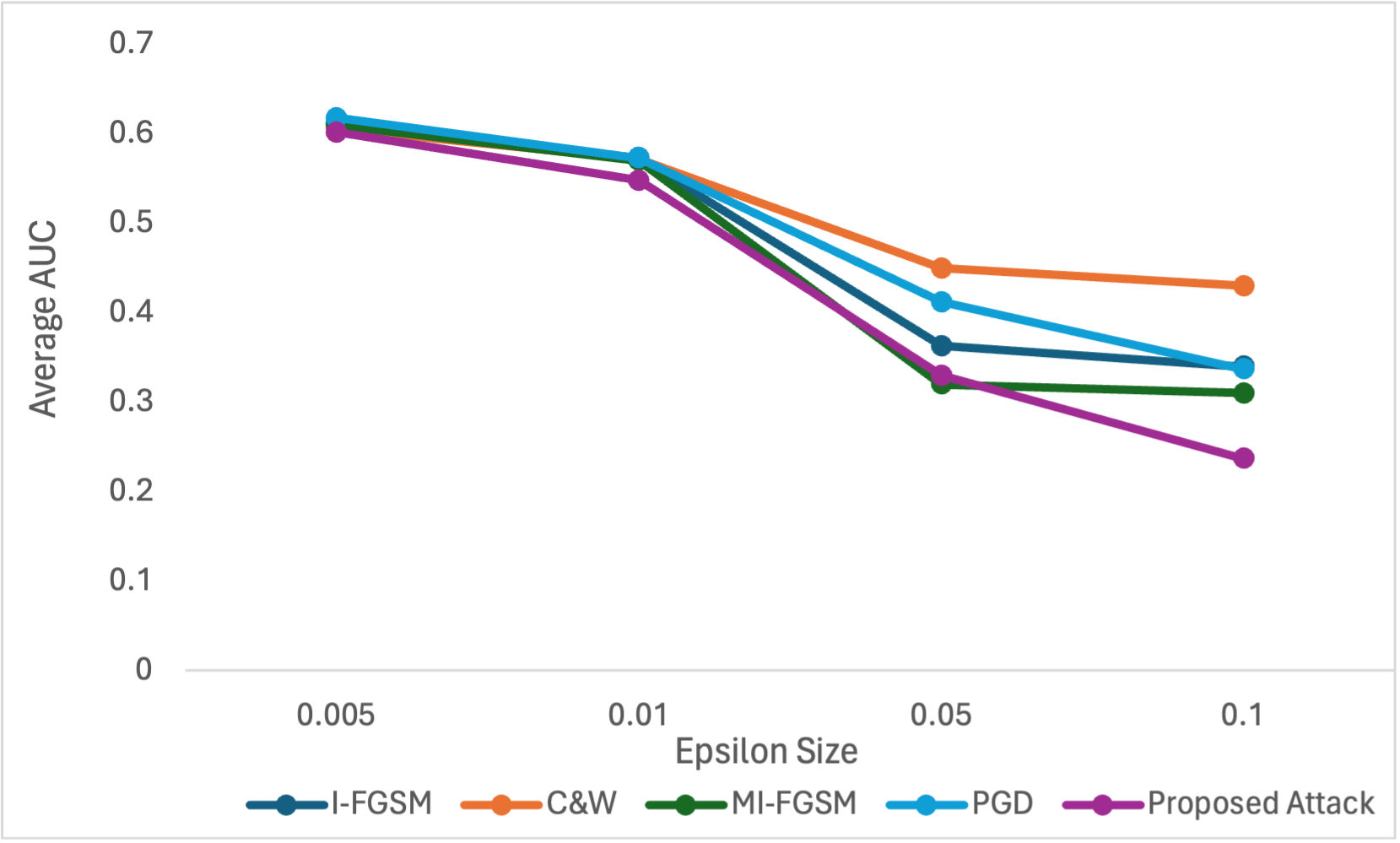}
    \caption{}
    \label{fig:sub2}
\end{subfigure}

\caption{\textbf{Parameter robustness analysis.} Target model performance is shown with respect to a range of values of the iteration number (A, B) and perturbation size (C, D). Adversarial training was applied in B and D. }
\label{fig:main}
\end{figure}
\section{Results}

As shown in Table 1, the Source and Target models show baseline AUCs of 0.670 and 0.704, respectively. The Target model has a higher AUC than the Source model, indicating the usefulness of the use of Prior exams for diagnosis. With adversarial attacks, all methods lowered the AUCs to a range of 0.205 to 0.531, showing all the attacks successfully fooled the model to give opposite (AUC<0.5) or random (AUC=0.5) diagnosis. Our proposed attack led to the lowest AUC of 0.205, outperforming all the other compared methods.

Table 1 last column shows the effects of after incorporating adversarial training to retrain the Target models. It shows that the Target model achieved an AUC of 0.685 when testing with clean data, which is slightly lower than the AUC (0.704) without adversarial training – this shows a trade-off between increased adversarial robustness and slightly-decreased performance on clean data. As can be seen, the Target model becomes more resilient to all the adversarial attacks as the AUC increased to higher values (range 0.548 to 0.634) compared to those (0.205 to 0.531) without adversarial training. This partially indicates the adversarial training is useful to mitigate the attacks to certain extent, but still, the model performance remains much lower than the normal performance (0.685), which means the attacks can still substantially fool the diagnosis model. Here, again, our proposed attack method achieved the best attacking effects.

The results also show that both the cross-entropy loss (FGSM, I-FGSM), PGD and distance-guided learning (Distance-guided FGSM, Distance-guided I-FGSM) can independently degrade model performance. The Distance Regularization-based methods perform better than the distance-guided methods, but are less effective than our proposed approach – this is potentially due to that the distance metric loss is not necessarily able to ensure that, in altering the relationship between the adversarial Current and Prior, the adversarial Current will breach the decision boundary, a task typically however can be influenced by cross-entropy loss. This suggests that our knowledge-guided sample selection method used to generate adversarial samples is a more effective method for attacks.

It should be noted that the generated adversarial samples are supposed to be able to fool the Source model (even though the real intention is to attack the Target model). This is verified in Table 1 from the low AUCs (0 means an opposite diagnosis of the entire cases) of the Source model. In addition, for the knowledge described for sample selection, our experiments also provided quantitative statistics supporting the validity of the distance knowledge: the average Euclidean distance between Prior and Current is 0.38 (smaller) and 0.52 (larger) on the normal and cancer patients, respectively.

Table 2 presents the performance of various attack methods using VGG as the backbone for the source model, while keeping the target models (including the adversarially trained model) unchanged. The results indicate that the overall attack performance patterns are consistent with those observed in Table 1, suggesting that our method is effective across different model architectures, including non-Transformer-based models.

From Fig. 3, our method consistently outperforms the other four methods (I-FGSM, C\&W, MI-FGSM, PGD) across a range of parameter values, both with and without adversarial training. Notably, as the epsilon value increases—indicating stronger adversarial perturbations—AUC values for all methods decrease, as expected.

\section{Conclusion}
In this study, we delved into the medical imaging AI model robustness against adversarial attacks on longitudinal models, with a particular focus on breast cancer diagnosis. Our research topic is novel because studies on adversarial attacks to longitudinal models are rare, yet such models are gaining popularity in medical applications. We proposed a novel attacking method that combines cross-entropy loss and knowledge-guided distance metric learning, showing much superior effects in terms of fooling the diagnosis model, and outperforming several compared state-of-the-art methods. Our method remained effective even after incorporating the defensing method of adversarial training. Future work includes further evaluation on different deep learning structures and developing effective defense methods. Our study highlights the importance and urgency of adversarially robust medical diagnosis models towards delivering safe AI to patient care.

\section{Acknowledgements}
This work was supported by a National Institutes of Health (NIH)/National Cancer Institute (NCI) grant (1R01CA218405), the grant 1R01EB032896 (and a Supplement grant 3R01EB032896-03S1) as part of the National Science Foundation (NSF)/NIH Smart Health and Biomedical Research in the Era of Artificial Intelligence and Advanced Data Science Program, a NSF grant (CICI: SIVD: \#2115082), an Amazon Machine Learning Research Award, and the University of Pittsburgh Momentum Funds (a scaling grant) for the Pittsburgh Center for AI Innovation in Medical Imaging. This work used Bridges-2 at Pittsburgh Supercomputing Center through allocation [MED200006] from the Advanced Cyberinfrastructure Coordination Ecosystem: Services \& Support (ACCESS) program, which is supported by NSF grants \#2138259, \#2138286, \#2138307, \#2137603, and \#2138296.
%
%


%
%

\end{document}